%% file: main.tex
\documentclass[10pt,twocolumn,letterpaper]{article}
\usepackage{algorithm,algorithmic,amsmath,amssymb,array,booktabs,colortbl,cvpr,epsfig,graphicx,rotating,setspace,subcaption,times,xcolor}
\usepackage[pagebackref=true,breaklinks=true, colorlinks,bookmarks=false]{hyperref}

\definecolor{Gray}{gray}{0.85}
\newcolumntype{g}{>{\columncolor{Gray}} c}
\newcolumntype{L}{>{\centering\arraybackslash}m{3cm}}

\cvprfinalcopy 

\def\cvprPaperID{139} 

\newcommand{\rot}[1]{\rotatebox{90}{\footnotesize{#1}}}

\newcommand{\tblheaderShort}{
		\rot{road} & \rot{sidewalk} & \rot{building} & \rot{wall} & \rot{fence} & 
							  \rot{pole} & \rot{t light} & \rot{t sign} & \rot{veg} & 
							  \rot{terrain} & \rot{sky} & \rot{person} & \rot{rider} & \rot{car} & 
							  \rot{truck} & \rot{bus} & \rot{train} & \rot{mbike} & \rot{bike} & mIoU}

\newcommand{\tblheaderSYNTHIA}{
	\rot{sky} & \rot{building} & \rot{road} & \rot{sidewalk} & \rot{fence} & 
	\rot{vegetation} & \rot{pole} & \rot{car} & \rot{t sign} & 
	\rot{pedestrian} & \rot{bicycle} & \rot{lanemarking} & \rot{t light} & mIoU}

\graphicspath{{figures/}}

\begin{document}

\title{FCNs in the Wild: Pixel-level Adversarial and Constraint-based Adaptation} 
%

\author{Judy Hoffman\\
CS Department\\
Stanford University\\
{\tt \small jhoffman@cs.stanford}
\and
Dequan Wang\\
EECS Department\\
UC Berkeley\\
{\tt\small dqwang@cs.berkeley}
\and
Fisher Yu\\
CS Department\\
Princeton University\\
{\tt\small i@yf.io}
\and
Trevor Darrell\\
EECS Department\\
UC Berkeley\\
{\tt\small trevor@cs.berkeley}
}

\maketitle

\input{abstract}

\input{introduction}
\input{relatedwork}
\input{method}
\input{experiments}
\input{conclusion}

\clearpage

\begin{spacing}{1.15}
\small
\bibliographystyle{ieee}
\bibliography{ref}
\end{spacing}
\end{document}

%% file: abstract.tex
\begin{abstract}
    Fully convolutional models for dense prediction have proven successful for a wide range of visual tasks. 
    Such models perform well in a supervised setting, but performance can be surprisingly poor under domain shifts that appear mild to a human observer. 
    For example, training on one city and testing on another in a different geographic region and/or weather condition may result in significantly degraded performance due to pixel-level distribution shift. 
    In this paper, we introduce the first domain adaptive semantic segmentation method, proposing an unsupervised adversarial approach to pixel prediction problems. 
    Our method consists of both global and category specific adaptation techniques.
    Global domain alignment is performed using a novel semantic segmentation network with fully convolutional domain adversarial learning.
    This initially adapted space then enables category specific adaptation through a generalization of constrained weak learning, with explicit transfer of the spatial layout from the source to the target domains. 
    Our approach outperforms baselines across  different settings on multiple large-scale datasets, including adapting across various real city environments, different synthetic sub-domains, from simulated to real environments, and on a novel large-scale dash-cam dataset.
\end{abstract}

%% file: introduction.tex
\section{Introduction}
\label{sec:introduction}

Semantic segmentation is a critical visual recognition task for a variety of applications ranging from autonomous agent tasks, such as robotic navigation and self-driving cars, to mapping and categorizing the natural world. 
As such, a significant amount of recent work has been introduced to tackle the supervised semantic segmentation problem using pixel-wise annotated images to train convolutional networks~\cite{long2015fully,chen2015semantic,noh2015learning,zheng2015conditional,liu2015semantic,dai2016instance,yu2016multi}.

\begin{figure}[t]
	\centering
	\includegraphics[width=1\linewidth]{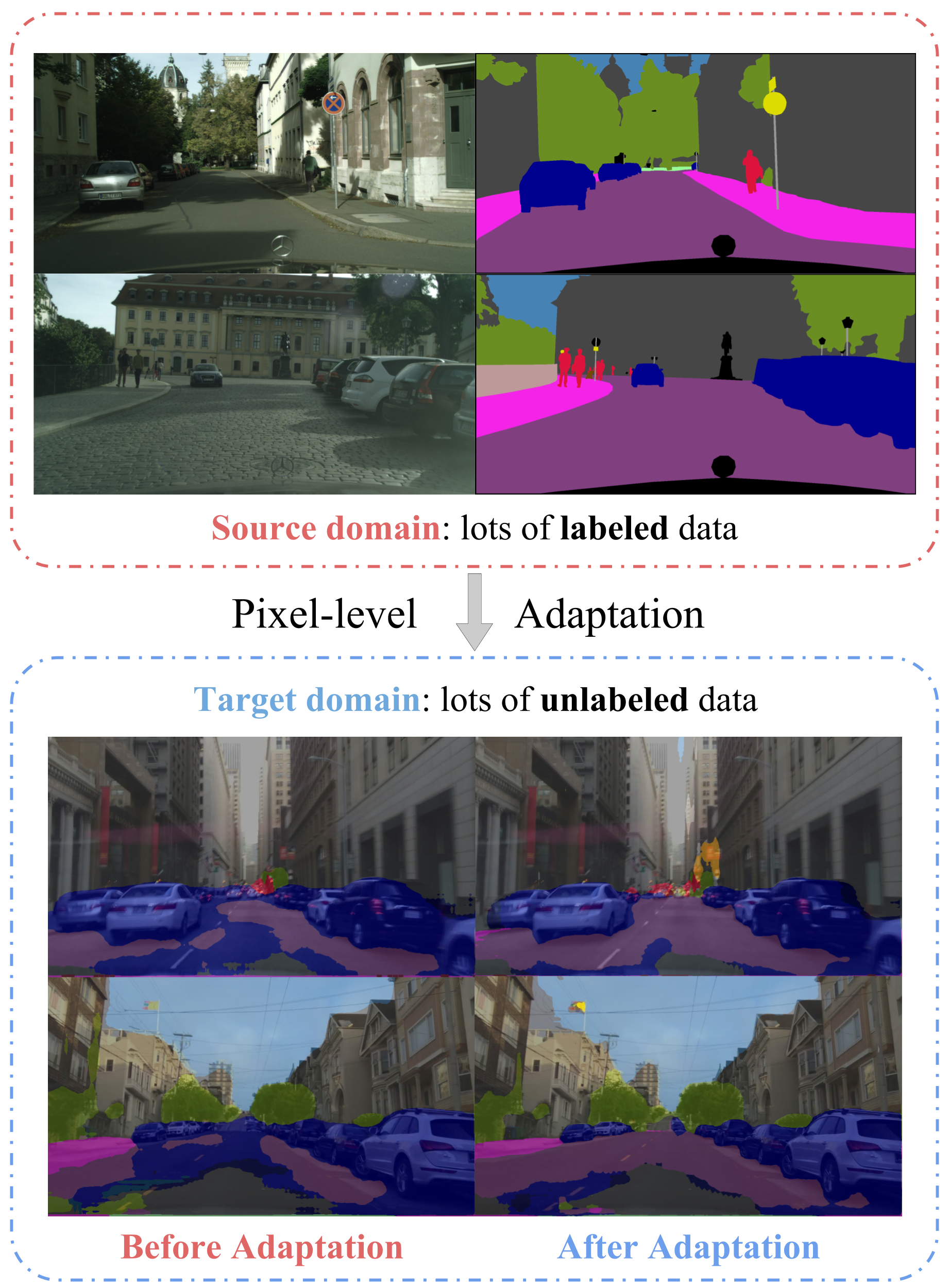}
	\caption{Unsupervised domain adaptation for pixel-level semantic segmentation.}
	\label{fig:challenge}
	\vspace{-5mm}
\end{figure}

While performance is improving for segmentation models trained and evaluated on the same data source, there has yet been limited research exploring the applicability of these models to new related domains. 
Many of the challenges faced when considering adapting between visual domains for classification, such as changes in appearance, lighting, and pose, are also present when considering adapting for semantic segmentation.
In addition, some new factors take on more prominence when considering recognition with localization tasks. 
In both classification and segmentation, the prevalence of classes may vary between different domains, but this variance can be more exaggerated with semantic segmentation applications as an individual object class may now appear many times within a single scene. For instance, semantic segmentation for self-driving applications will focus on outdoor street scenes with objects of varying sizes, whose distribution may vary between cities or driving routes; in addition appearance statistics can vary considerably when, e.g., adapting a person recognition model trained only using indoor scene images. 
Moreover, pixel-wise annotations are expensive and tedious to collect, making it particularly appealing to learn to share and transfer information between related settings. 

In this work, we propose the first unsupervised domain adaptation method for transferring semantic segmentation FCNs across image domains. 
A second contribution of our approach is the combination of global and local alignment methods, using global and category specific adaptation techniques that are themselves individually innovative contributions. 
We align the global statistics of our source and target data using a convolutional domain adversarial training technique, using a novel extension of previous image-level classification approaches~\cite{tzeng2015simultaneous,ganin2015unsupervised,ganin2016domain}.
Given a domain aligned representation space, we introduce a generalizable constrained multiple instance loss function, which expands on weak label learning~\cite{pathak2015fully,pathak2015constrained,pinheiro2015image,papandreou2015weakly,hong2015decoupled}, but can be applied to the target domain without any extra annotations and explicitly transfers category layout information from a labeled source dataset.


We evaluate our approach using multiple 
large scale datasets.
We first make use of recently released synthetic drive-cam data from both the GTA5~\cite{richter2016playing} and SYNTHIA~\cite{ros2016synthia} datasets, in order to examine a large adaptation shift from simulated to the real images available in CityScapes~\cite{cordts2016cityscapes}. Next, we explore the domain shift of cross season adaptation within the SYNTHIA dataset. We then focus on adaptation across cities in the real world. We perform a detailed quantitative analysis of cross-city adaptation within the CityScapes dataset. 

A final contribution of our paper is the  introduction of a new unconstrained drive-cam dataset for semantic segmentation, Berkeley Deep Driving Segmentation (BDDS). 
Below we demonstrate initial qualitative adaptation results from Cityscapes cities to the cities in BDDS.
Across all of these studies, we show that our adaptation algorithm improves the target semantic segmentation performance without any target annotations.

%% file: relatedwork.tex
\section{Related Work}
\label{sec:relatedwork}

\paragraph{Semantic Segmentation}  
Semantic segmentation is a key computer vision task and has been studied in a plethora of publications. 
Following the success of large-scale image classification, most current semantic segmentation models use some convolutional network architecture \cite{farabet2013learning,hariharan2014simultaneous} with many recent approaches using fully convolutional networks (FCNs) \cite{long2015fully} to map the input RGB space to a semantic pixel space. 
These models are compelling because they allow a direct end-to-end function that can be trained using back propagation. 
The original FCN formulation has since been improved using dilated convolution \cite{yu2016multi} and post-processing techniques, such as Markov/conditional random fields \cite{chen2015semantic,liu2015semantic,zheng2015conditional}. 

Motivated by the high cost of collecting pixel level supervision, a related body of work has explored using weak labels (typically image-level tags defining presence / absence of each class), to improve semantic segmentation performance. 
Pathak \etal \cite{pathak2015fully} and Pinheiro \etal \cite{pinheiro2015image} modeled this problem as multiple instance learning (MIL) and reinforce confident predictions during the learning process. 
An improved method was suggested by \cite{papandreou2015weakly} who use an EM algorithm to better model global properties of the image segments. 
This work was in turn generalized by Pathak \etal who proposed a Constrained CNN which is able to model any linear constraints on the label space (\ie presence / absence, percent cover) \cite{pathak2015constrained}. 
In another recent paper~\cite{hong2016learning}, Hong \etal used auxiliary segmentation to generalize semantic segmentations to categories where only weak label information was available.

From a domain adaptation perspective, these methods all assume that weak labels are present during training time for both source domain and target domain. 
In this work, we consider a related, but different learning scenario: strong supervision is available in the source domain, but that no supervision is available in the target domain. 
\vspace{-3mm}
\paragraph{Domain Adaptation}  
Domain adaptation in computer vision has focused largely on image classification, with much work dedicated to generalizing across the domain shift between stock photographs of objects and the same objects photographed in the world \cite{saenko2010adapting,kulis2011you,gong2012geodesic}. 
Recent work includes \cite{tzeng2015simultaneous,ganin2015unsupervised,ganin2016domain} which all learn a feature representation which encourages maximal confusion between the two domains. 
Other work aims to align the features \cite{long2015learning,long2016unsupervised} by minimizing the distance between their distributions in the two domains. 
Based on Generative Adversarial Network \cite{goodfellow2014generative}, Liu \etal proposed coupled generative adversarial network to learn a joint distribution of images from both source and target datasets \cite{liu2016coupled}.

Much less attention has been given to other important computer vision tasks such as detection and segmentation. 
In detection, Hoffman \etal proposed a domain adaptation system by explicitly modeling the representation shift between classification and detection models \cite{hoffman2014lsda} along with a follow-up work which incorporated per-category adaptation using multiple instance learning \cite{hoffman2015detector}. 
The detection models were later converted into FCNs for evaluating semantic segmentation performance \cite{hoffman2016large}, but this work did not propose any segmentation specific adaptation approach. 
So far as we know, our method is the first to introduce domain adaptation techniques for semantic segmentation models.


%% file: method.tex
\newcommand{\src}{\mathcal{S}}
\newcommand{\tgt}{\mathcal{T}}
\newcommand{\img}{{I}}
\newcommand{\anno}{L}
\newcommand{\net}{\phi}
\newcommand{\srcNet}{\phi_{\src}}
\newcommand{\tgtNet}{\phi_{\tgt}}
\newcommand{\srcImg}{\img_{\src}}
\newcommand{\srcLabel}{\anno_{\src}}
\newcommand{\tgtImg}{\img_{\tgt}}
\newcommand{\tgtLabel}{\anno_{\tgt}}

\newcommand{\loss}{\mathcal{L}}
\newcommand{\daLoss}{\loss_{da}}
\newcommand{\segLoss}{\loss_{seg}}
\newcommand{\miLoss}{\loss_{mi}}
\newcommand{\srcLabelStat}{\mathcal{P}_{\srcLabel}}

\section{Fully Convolutional Adaptation Models}
\begin{figure}[t]
	\begin{center}
		\includegraphics[width=1\linewidth]{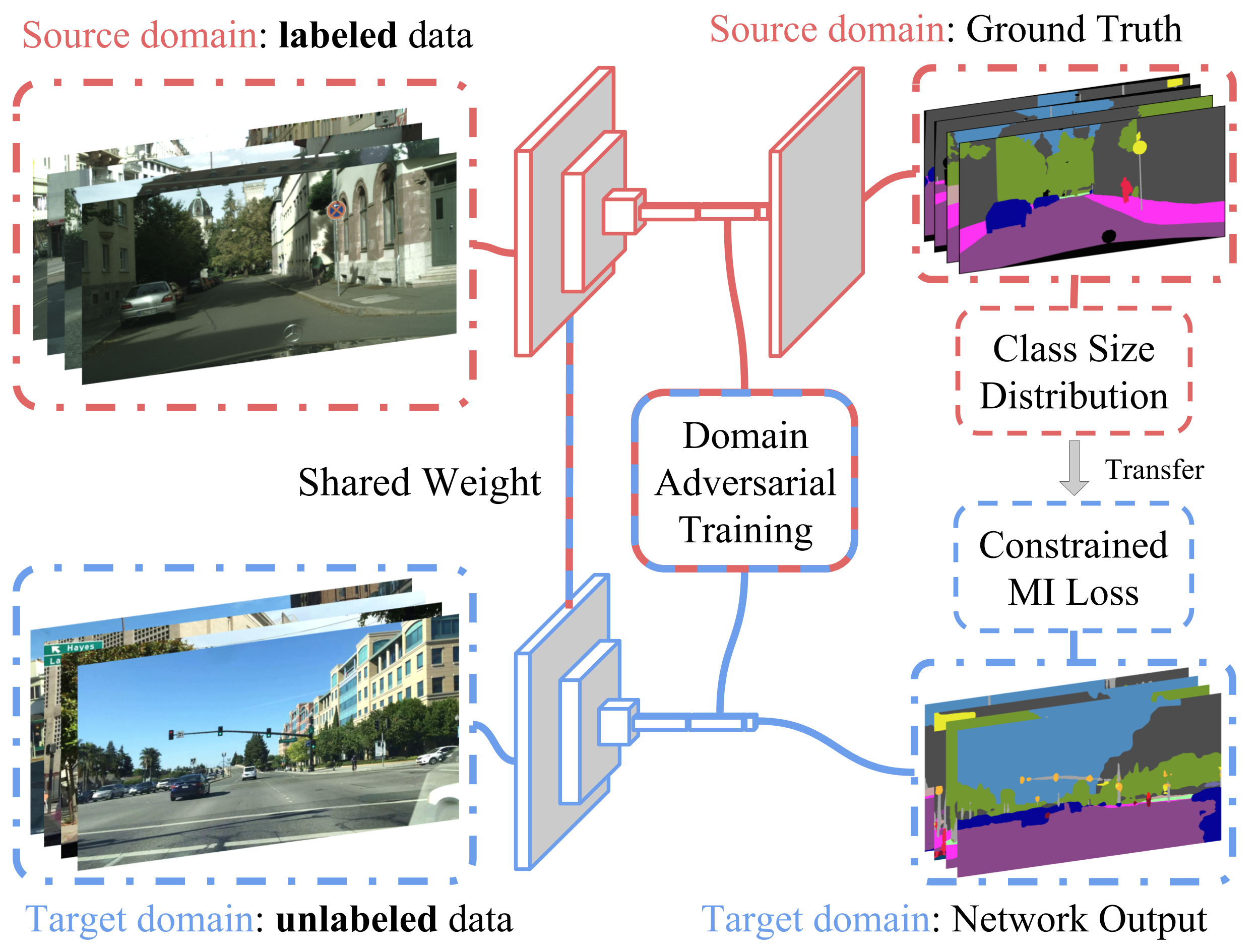}
	\end{center}
    \vspace{-3mm}
	\caption{Overview of our pixel-level adversarial and constraint-based adaptation.}
	\vspace{-3mm}
    \label{fig:framework}
\end{figure}

In this section, we describe our adaptation algorithm for semantic segmentation using fully convolutional networks (FCNs) across domains which share a common label space. Without loss of generality, our method can be applied to other segmentation models, though we focus here on FCNs due to their broad impact. We consider having access to a source domain, $\src$, with both images, $\srcImg$, and labels, $\srcLabel$. 
We train a source only model for semantic segmentation which produces a pixel-wise per-category score map $\srcNet(\srcImg)$.

Our goal is to learn a semantic segmentation model which is adapted for use on the unlabeled target domain, $\tgt$, with images, $\tgtImg$, but no annotations. We denote the parameters of such as network as $\tgtNet(\cdot)$. If there is no domain shift between the source and target domains then one could simply apply the source model directly to the target with no need for an adaptive approach. 
However, there is commonly a difference between the distribution of the source labeled domain and the target test domain. 

Therefore, we present an \textit{unsupervised adaptation} approach. We begin by noting that there are two main opportunities for domain shift. First, global changes may occur between the two domains resulting in a marginal distribution shift of corresponding feature space. This may occur between any two different domains, but will be most distinct in large shifts between very distinct domains, such as adapting between simulated and real domains.
The second main shift occurs due to category specific parameter changes. This may result from individual categories having specific biases in the two domains. For example, when adapting between two different cities the distribution of cars and the appearance of signs may change.


We propose an unsupervised domain adaptation framework for adapting semantic segmentation models which directly tackles both the need for minimizing the global and the category specific shifts. For our model, we first make the necessary assumption that the source and target domains share the same label space and that the source model achieves performance greater than chance on the target domain. Then, we introduce two new semantic segmentation loss objectives, one to minimize the global distribution distance, which operates over both source and target images, $\daLoss(\srcImg, \tgtImg)$.  Another to adapt the category specific parameters using target images and transferring label statistics from the source domain $\srcLabelStat$, $\miLoss(\tgtImg, \srcLabelStat)$. Finally, to ensure that we do not diverge too far from the source solution, which is known to be effective for the final semantic segmentation task, we continue to optimize the standard supervised segmentation objective on the source domain, $\segLoss(\srcImg, \srcLabel)$. Together, our adaptive learning approach is to optimize the following joint objective: 
\begin{eqnarray}
	\loss(\srcImg, \srcLabel, \tgtImg) &=& \segLoss(\srcImg, \srcLabel) \\
	&& + \daLoss(\srcImg, \tgtImg) + \miLoss(\tgtImg, \srcLabelStat) \nonumber
\end{eqnarray}

We illustrate overall adaptation framework in Figure~\ref{fig:framework}. Source domain data is used to update the standard supervised loss objective, trained using the source pixel-wise annotations. Both source and target data are used without any category annotations within fully-convolutional domain adversarial training to minimize the global distance of feature space between the two domains. Finally, category specific updates using a constrained pixel-wise multiple instance learning objective is performed on the target images, with source category statistics used to determine the constraints.

Note, our approach may be generally applied to any 
FCN-based semantic segmentation framework. For our experiments, we use the recently proposed front-end dilated fully convolutional network~\cite{yu2016multi}, based on 16 layers VGGNet~\cite{simonyan2015very}, as our base model. There are 16 convolutional layers, where the last three convolutional layer converted from fully connected layers, called $fc_6, fc_7, fc_8$, followed by $8$ times bilinear up-sample layer to produce segmentation in the same resolution as input image.

\subsection{Global Domain Alignment}
\label{sec:global}

We begin by describing in more detail our global domain alignment objective, $\daLoss(\srcImg, \tgtImg)$. 
Recall, that we seek to minimize the domain shift between representations of the source and target data. 
A recent line of research has shown that the domain discrepancy distance may be minimized through an adversarial learning procedure, whereby simultaneously a domain classifier is trained to best distinguish the source and target distributions and the representation space is updated according to the inverse objective~\cite{tzeng2015simultaneous,chen2016synthesizing,ganin2016domain}. The approaches heretofore have been introduced for classification models where each individual instance in the domain corresponds exactly to an image. 

Here, we propose a new domain adversarial learning objective which may be applied for pixel-wise approaches to aid in learning domain invariant representations for semantic segmentation models. The first question to answer is what should comprise an instance within the dense prediction framework. Since recognition is sought at the pixel level alignment of full image representations will marginalize out too much distribution information limiting the alignment capability of the adversarial learning approach. 

Instead, we consider the region corresponding to the natural receptive field of each spatial unit in the final representation layer (\eg $fc_7$),  as individual instances. 
In doing so, we directly supply our adversarial training procedure with the same information which is used to do final pixel prediction. Therefore, this provides a more meaningful view of the overall source and target pixel-space representation distribution distance which needs to be minimize. 
%
%
 
Let $\net_{\ell-1}(\theta, \img)$ denote the output of the last layer before pixel prediction according to network parameters, $\theta$. 
Then, our domain adversarial loss, $\daLoss(\srcImg, \tgtImg)$ consists of alternating minimization objectives. One concerning the parameters of the representation space, $\theta$, under which we would like to minimize the observed source and target distance, $\min d(\net_{\ell-1}(\theta,\srcImg), \net_{\ell-1}(\theta,\tgtImg)$, for a given distance function, $d(\cdot)$. The second concerning estimating the distance function through training a domain classifier to distinguish instances of the source and target domains. Let us denote the domain classifier parameters as $\theta_D$. 
We then seek to learn a domain classifier to recognize the difference between source and target regions and use that classifier to guide the distance minimization of the source and target representations. 
 
%

Let $\sigma(\cdot)$ denote the softmax function and let the domain classifier predictions be indicated as $p_{\theta_D}(x) = \sigma(\net(\theta_D, x)))$. Assuming the output of layer $\ell-1$ has $H\times W$ spatial units, then we can define the domain classifier loss, $\loss_D$, as follows:
\begin{eqnarray}
	\loss_D &=& -\sum_{\srcImg \in \src} \sum_{h \in H} \sum_{w \in W} \log(p_{\theta_D}(R^{\src}_{hw})) \\
			&& - \sum_{\tgtImg \in \tgt} \sum_{h \in H} \sum_{w\in W} \log(1-p_{\theta_D}(R^{\tgt}_{hw}))
\end{eqnarray}
where $R^{\src}_{hw} = \net_{\ell-1}(\theta, \srcImg)_{hw}$ and $R^{\tgt}_{hw} = \net_{\ell-1}(\theta, \tgtImg)_{hw}$ denote the source and target representation of each units, respectively.

For convenience let us also define the inverse domain loss, $\loss_{Dinv}$ as follows:
\begin{eqnarray}
	\loss_{Dinv} &=& - \sum_{\srcImg \in \src} \sum_{h \in H} \sum_{w \in W} \log(1-p_{\theta_D}(R^{\src}_{hw})) \\
			&& - \sum_{\tgtImg \in \tgt} \sum_{h \in H} \sum_{w\in W} \log(p_{\theta_D}(R^{\tgt}_{hw}))
\end{eqnarray}
Finally, with these definitions, we may now describe the alternating minimization procedure.
 \begin{eqnarray}
	\min_{\theta_D} & \loss_D \label{eq:dom_loss} \\
	\min_{\theta} & \frac{1}{2} \left[\loss_D + \loss_{Dinv} \right] \label{eq:dom_dist}
\end{eqnarray}
Optimizing these two objectives iteratively amounts to learning the best possible domain classifier for relevant image regions (Eq~\eqref{eq:dom_loss}) and then using the loss of that domain classifier to inform the training of the image representations so as to minimize the distance between the source and target domains (Eq~\eqref{eq:dom_dist}).

\subsection{Category Specific Adaptation}
\label{sec:category}

Given our representation which has minimized the global domain distribution distance through our fully convolutional adversarial training objective, the next step is to further adapt our source model through modifying the category specific network parameters. In order to do this, we draw upon recent weak learning literature~\cite{pathak2015constrained,pathak2015fully}, which introduced a fully convolutional constrained multiple instance learning objective. This work used size and existence constraints to produce a predicted target labeling to use for further training. We present the novel application of such approaches for domain adaptation and generalize the technique for use in our unlabeled setting. 

First, we consider new constraints which are useful for our pixel-wise unsupervised adaptation problem. In particular, we begin by computing per image labeling statistics in the source domain, $\srcLabelStat$. Specifically, for each source image which contains class $c$, we compute the percentage of image pixels which have a ground truth label corresponding to this class. We can then compute a histogram over these percentages and denote the lower 10\% boundary as, $\alpha_c$, the average value as $\delta_c$, and the upper 10\% as $\gamma_c$. We may then use this distribution to inform our target domain size constraints, thereby explicitly transferring scene layout information from the source to the target domain. For example, in a driving scenario, often the road occupies a large portion of the image while street signs occupy relatively little image real estate. This information is critical to the constrained multiple instance learning procedure. In contrast, prior work used a single size threshold across classes known to be in the image.

We begin by presenting our constrained multiple instance loss for the case where image-level labels are known. Thus, for a given target image for which a certain class $c$ is present, we impose the following constraints on the output prediction map, $p = \text{arg}\max \net(\theta, \tgtImg)$.

\begin{equation}
	\delta_c \le \sum_{h,w} p_{hw}(c) \le \gamma_c
\end{equation}

Thus, our constraint encourages pixels to be assigned to class $c$ such that the percentage of the image labeled with class $c$ is within the expected range observed in the source domain.
Practically, we optimize this objective with lower bound slack to allow for outlier cases where $c$ simply occupies less of the image than is average in the source domain. However, we do not allow slack on the upper bound constraint as it is important that no single class occupies too much of any given image.
Notice that our updated constraint is general and can be equivalently applied to all classes regardless if they correspond with traditional object notion (\eg bikes or people) or stuff notion (\eg sky or vegetation). 

Given this constraint we may now optimize for a new class prediction space to use for future learning. For the specific optimization details we refer the reader to Pathak \etal~\cite{pathak2015constrained}. We provide one important modification. As we seek to optimize over both object and stuff categories, we note that the relative number of pixels devoted to each may vary significantly which could cause the model to diverge, over-fitting to those classes which are highly represented in the images. Instead, we use a simple size constraint that if the lower 10\% of the source class distribution, $\alpha_c$, is greater than 0.1, then we down-weight the gradients due to these classes by a factor of 0.1. This is re-weighting approach can be viewed as a re-sampling of the classes so as to come closer to a balanced set, allowing the relatively small classes potential to inform the learning objective.

While the approach described above describes a generalized constrained multiple instance objective, it relies on known image-level labels. Since we lack such information in our unsupervised adaptation setting, we now describe our procedure for predicting image level labels. Thus, our complete approach can be described as first predicting image-level labels and then optimizing for pixel predictions which satisfy the source transferred class size constraints. 

In contrast to weakly-supervised settings, we do not learn a segmentation model from scratch with known image-level annotations. Instead, we have access to a fully supervised source dataset and use domain transferred constraints to facilitate transfer to an unsupervised target domain. Thus we both have a stronger initial model with a fully supervised using pixel-level annotations from source domain and are additionally able to regularize the learning procedure by training with weak label loss on target domain.
Again, given a target image, $\tgtImg$, we compute the current output class prediction map, $p = \arg\max\net(\theta, \tgtImg)$. For each class we compute the percentage of pixels assigned to that class in our current prediction, $d_c = \frac{1}{H\cdot W}\sum_{h \in H} \sum_{w\in W} (p_{hw} = c)$. Finally, we assign an image-level label to class $c$ if $d_c > 0.1 * \alpha_c$, meaning if we currently label at least as many pixels as 10\% of the expected number for a true class appearing in the image.

%% file: experiments.tex
\label{sec:experiments}
\section{Experiments}
In this section, we report our experimental results on three different domain adaptation tasks: $cities\rightarrow cities$, $season\rightarrow season$, and $synthetic\rightarrow real$, studied across four different datasets. We analyze both our overall adaptation approach as well as the sub-components to verify that both our global and category specific alignment offer meaningful contributions.

For all experiments we use the front-end dilated fully convolutional network~\cite{yu2016multi} as both the initialization for our method and as the baseline model for comparison. All code and models are trained and evaluated in the  Caffe~\cite{jia2014caffe} framework and will be made available before camera-ready.

\input{table_synthetic2real}

For fair comparison, we use the Intersection over Union (IoU) evaluation metric for all experiments. For $cities\rightarrow cities$ and $synthetic\rightarrow real$ tasks, we followed the evaluation protocol of \cite{cordts2016cityscapes} and train our models with $19$ semantic labels of Cityscapes. For $season\rightarrow season$ task, we use $13$ semantic labels of SYNTHIA instead. 

\subsection{Datasets}

\textbf{Cityscapes} contains 34 categories in high resolution, $2048\times1024$. The whole dataset is divided into three parts: $2,975$ training samples, $500$ validation samples and $1,525$ test samples. The split of this dataset is city-level, which covers individual European cities in different geographic and population distribution. 


\textbf{SYNTHIA} contains 13 classes with different scenarios and sub-conditions. As for $season\rightarrow season$ task, we regard SYNTHIA-VIDEO-SEQUENCES as play ground. There are $7$ sequences, covering different scenarios (highway, roundabout, mountain path, New York City, Old European Town) with several sub-sequences, such as seasons(Spring, Summer, Fall, Winter), weathers(Rain, Soft-Rain, Fog), and illuminations(Sunset, Dawn, Night). These frames are captured by $8$ RGB cameras forming a binocular $360^\circ$ visual field. In order to minimize the impact of viewpoint, we only pick up the dashcam-like frames for all the time.  As for $synthetic\rightarrow real$ task, we take SYNTHIA-RAND-CITYSCAPES, providing $9,000$ random images from all the sequences with Cityscape-compatible annotations, as source domain data. 

\textbf{GTA5} contains $24,966$ high quality labeled frames from realistic open-world computer games, Grand Theft Auto V (GTA5). Each frame, with high resolution $1914\times1052$, is generated from fictional city of Los Santos, based on Los Angeles in Southern California. We take the whole dataset with labels compatible to Cityscapes categories for $synthetic\rightarrow real$ adaptation.

\textbf{BDDS} contains thousands of dense annotated dash-cam video frames and hundreds of thousands of unlabeled frames. Each sample, with high resolution $1280\times720$, provides $34$ categories compatible to Cityscapes label space. The majority of our data comes from New York and San Francisco, which are the representative of eastern and western coasts. Different from the other existing driving datasets, this dataset covers diverse driving scenarios under different conditions, such as urban street view at night, highway scene in rain and so on, providing challenging domain adaptation settings.

\subsection{Quantitative and Qualitative Results}

We broadly study three types of shifts. First we study a large distribution shift, as seen when adapting from simulated to real imagery. Next, we study a medium sized shift, through adaptation across season patterns observed within the SYNTHIA dataset. Finally, we explore situations of relatively smaller domain shift, though exploring adaptation between different cities within the CityScapes dataset.

\subsubsection{Large Shift: Synthetic to Real Adaptation}
We begin the evaluation of our method by studying the large domain shift of adapting between simulated driving data and real world drive-cam data. Table~\ref{table:sim2real} shows semantic segmentation performance for the shift between GTA5 to CityScapes and between SYNTHIA to CityScapes. This illustrates that even with this large domain difference our unsupervised adaptation solution is capable of improving the performance of the source dilation model. Notice that for this larger shift setting, such as GTA5$\rightarrow$Cityscapes, the domain adversarial training contributes $4.4\%$ raw and $\scriptsize{\sim} 20\%$ relative percentage mIoU improvement and multiple instance loss contributes yet another $1.6\%$ raw and $\scriptsize{\sim} 6\%$ relative percentage mIoU improvement. 
As for SYNTHIA$\rightarrow$Cityscapes, our method also offers a measurable improvement.

\input{fig_synthia_fall2winter}
\input{table_season2season}
\input{table_city2city}
\subsubsection{Medium Shift: Cross Seasons Adaptation}
As our next experiment, we seek to analyze adaptation across season patterns. To this end, we use the SYNTHIA dataset which has synthetic images available along with season annotations. We first produce one domain per each of the season labels available: Summer, Fall and Winter. We then perform adaptation across each of the $6$ shifts and report the performance of our method in comparison to the source dilation model in Table~\ref{table:season2season}. On average we get $\scriptsize{\sim} 3$ percentage mIoU improvement for $season\rightarrow season$ adaptation and find that for 12/13 object categories our adaptation method provides higher mIoU. The one class we saw no improvement after adaptation is for \texttt{car}. We presume this results from the fact that cars have little or no appearance difference across seasons in this synthetic dataset. For example, consider the qualitative results shown in Figure~\ref{fig:season2season} for the shift of fall to winter. While the roads and side-walks have been rendered in a white to simulate snow in winter, the cars are rendered in the same appearance as in fall. In fact some of the largest performance improvements we saw from our method we in categories like \texttt{road} in the shift of fall to winter, and our method is able to overcome this large appearance shift. 

\input{fig_cityscapes2nexar}

\subsubsection{Small Shift: Cross City Adaptation}
For our third quantitative experiment we move towards studying cross city adaptation within the CityScapes dataset. 
In Table~\ref{table:city2city} we report performance on the task of adapting between the labeled cities in the CityScapes \textit{train} to the unlabeled cities in either the Cityscapes \textit{val}. The top row shows the performance of the dilation frontend model~\cite{yu2016multi}.
We report performance after only global alignment through domain adversarial training (indicated as Our method (GA only)) and after the category specific alignment with the constrained multiple instance loss (indicated as Our method (GA+CA)). 
We note that for this adaptation experiment the majority of the improvement from our method is as a result of the domain adversarial training ($3.6$ percentage mIoU) whereas after category specific alignment only offers a noticeable improvement on the categories of \texttt{traffic light}, \texttt{rider} and \texttt{train}. 
One reason could be that the domain shift between $train$ and $val$ mainly results from a change in the global appearance, due to the difference in city, whereas the specific category appearance may not change that significantly. 
Since performance on this within dataset adaptation is already quite high, the primary improvements arise from producing more consistent within object segmentations.

\subsection{BDDS Adaptation}
Finally, we analyze another real world $cities\rightarrow cities$ adaptation using our new large scale driving image dataset BDDS.
To understand this difficulty and evaluate our methods more extensively, we create a new image dataset based on dash cam videos. 
Although CityScapes covers various cities in Germany and neighboring countries, we observe that the cities in the other places have different visual appearance and street layout. They may post serious challenges to the models learned from CityScapes.
Up to now, we have collected more than $100,000$ images covering outdoor scenes at different time and places. Based on the current annotation progress, there would $5,000 \scriptsize{\sim} 10,000$ images with fine segmentation annotation before CVPR 2017. We aim for $10,000 \scriptsize{\sim} 20,000$ finely segmented street scene images eventually. 

We take $\scriptsize{\sim} 60,000$ images in area of San Francisco from BDDS and study how well we can adapt the model learned on Cityscapes to San Francisco. Because our methods don't require labels in the target domain, we can use all the new images in our training the adaptation. 
Some results are shown in Figure~\ref{fig:city2nexar}. From these qualitative results, we observe that there is a significant segmentation quality drop when a model trained on Cityscapes is used in BDDS directly. It usually appears as noisy segmentation or wrong context. After adaptation, the segmentation results usually become much cleaner. We expect to conduct extensive quantitative evaluation when annotations are ready.

%% file: table_synthetic2real.tex
\begin{table*}[ht]
	\small
	\setlength{\tabcolsep}{2.5pt}
	\begin{center}
		\begin{tabular}{l | ccccccccccccccccccc  g}
			\toprule
			\multicolumn{21}{c}{GTA5 $\rightarrow$ Cityscapes }\\
			\midrule
			Method & \tblheaderShort \\
			\midrule
			Dialation Frontend~\cite{yu2016multi} & 31.9 & 18.9 & 47.7 & 7.4 & 3.1 & \bf 16.0 & 10.4 & 1.0 & 76.5 & 13.0 & 58.9 & 36.0 & 1.0 & 67.1 & \bf 9.5 & 3.7 & 0.0 & 0.0 & 0.0 & 21.1 \\ 
			Our Method (GA only) & 67.4 & 29.2 & \bf 64.9 & \bf 15.6 & \bf 8.4 & 12.4 & 9.8 &\bf 2.7 & 74.1 & 12.8 & \bf 66.8 & 38.1 & 2.3 & 63.0 & 9.4 & 5.1 & 0.0 & 3.5 & 0.0 & 25.5 \\ 
			Our Method (GA + CA) & \bf 70.4&\bf 32.4& 62.1& 14.9& 5.4& 10.9& \bf 14.2& \bf 2.7& \bf 79.2& \bf 21.3&  64.6& \bf 44.1& \bf 4.2& \bf 70.4& 8.0& \bf 7.3& 0.0& \bf 3.5& 0.0& \bf 27.1\\
			\bottomrule
			\multicolumn{21}{c}{  }\\
			\toprule
			\multicolumn{21}{c}{SYNTHIA $\rightarrow$ Cityscapes }\\
			\midrule
			Dialation Frontend~\cite{yu2016multi} & 6.4 & 17.7 & 29.7 & 1.2 & 0.0 & 15.1 & 0.0 & 7.2 & 30.3 & 0.0 & 66.8 & 51.1 & 1.5 &  47.3 & 0.0 & 3.9 & 0.0 & 0.1 & 0.0 & 14.7\\ 
			Our Method (GA only) & \bf 11.5 & 18.3 & \bf 33.3 & \bf 6.1 & 0.0 & \bf 23.1 & 0.0 & 11.2 & \bf 43.6 & 0.0 & \bf 70.5 & 45.5 & 1.3 & 45.1 & 0.0 & \bf 4.6 & 0.0 & 0.1 &  0.5 & 16.6 \\ 
			Our Method (GA + CA) & \bf 11.5& 	\bf 19.6& 	30.8& 	4.4& 	0.0& 	20.3& 	0.1& 	\bf 11.7& 	42.3& 	0.0& 	68.7& 	\bf 51.2& 	\bf 3.8& 	\bf 54.0& 	0.0& 	3.2& 	0.0& 	\bf 0.2& 	\bf 0.6& 	\bf 17.0\\
			\bottomrule
		\end{tabular}
	\end{center}
	\vspace{-2mm}
	\caption{\textbf{Adaptation from synthetic to real.} We study the performance using GTA5 and SYNTHIA as source labeled training data adapted and Cityscapes \textit{train} as an unlabeled target domain, while evaluating our adaptation algorithm on Cityscapes \textit{val}. Meanwhile, we show an ablation of the components of our method and how each contributes to the overall performance of our approach. Here GA represents global domain alignment and CA indicates category specific adaptation.}
	\label{table:sim2real}
	\vspace{-4mm}
\end{table*}

%% file: fig_synthia_fall2winter.tex
\begin{figure*}[t!]
	\centering
	\begin{subfigure}[t]{0.196\linewidth}
		\centering
		\includegraphics[width=1\linewidth]{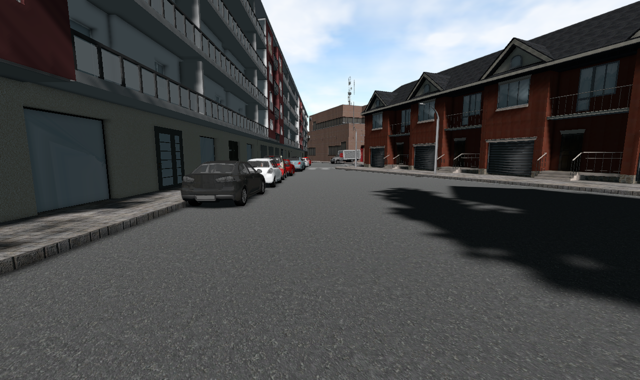}
	\end{subfigure}
	\hfill
	\begin{subfigure}[t]{0.196\linewidth}
		\centering
		\includegraphics[width=1\linewidth]{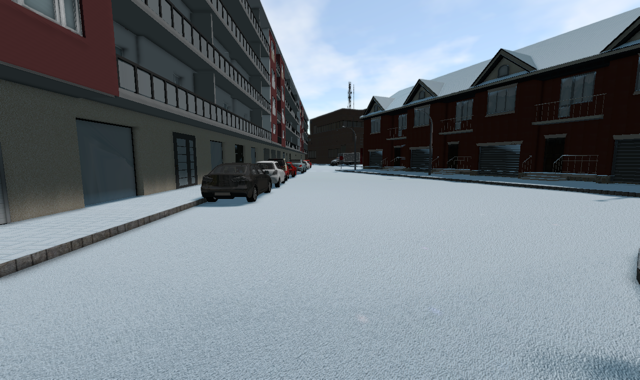}
	\end{subfigure}
	\hfill
	\begin{subfigure}[t]{0.196\linewidth}
		\centering
		\includegraphics[width=1\linewidth]{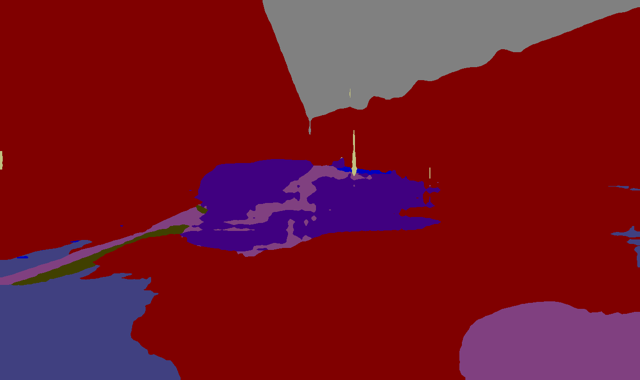}
	\end{subfigure}
	\hfill
	\begin{subfigure}[t]{0.196\linewidth}
		\centering
		\includegraphics[width=1\linewidth]{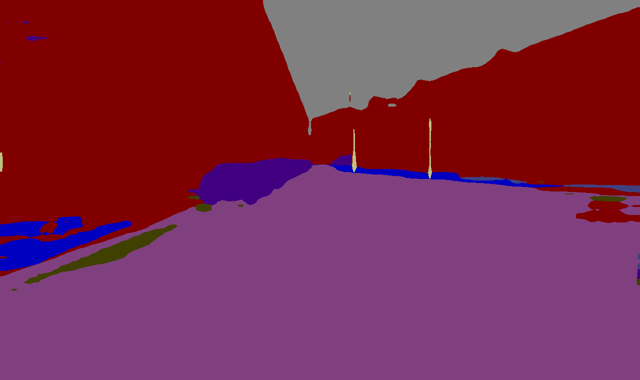}
	\end{subfigure}
	\hfill
	\begin{subfigure}[t]{0.196\linewidth}
		\centering
		\includegraphics[width=1\linewidth]{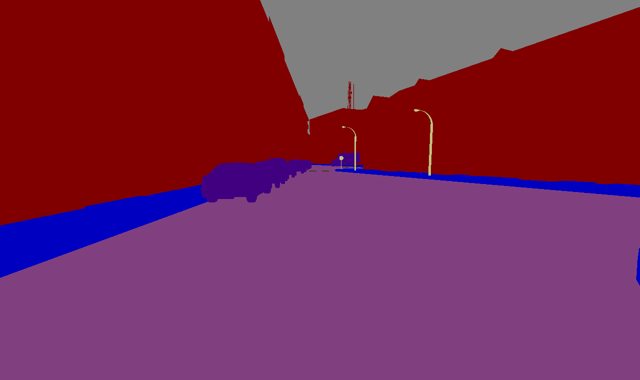}
	\end{subfigure}
	\hfill
	\\
	\centering
	\begin{subfigure}[t]{0.196\linewidth}
		\centering
		\includegraphics[width=1\linewidth]{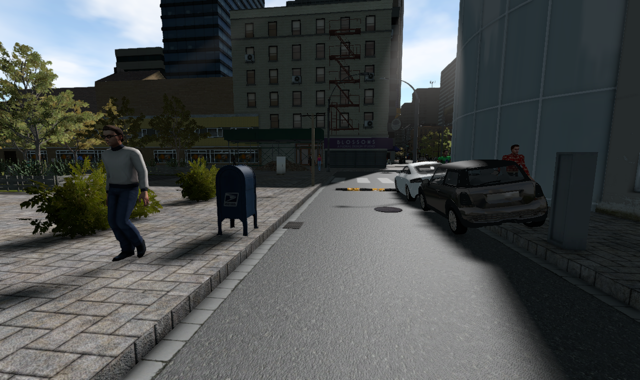}
	\end{subfigure}
	\hfill
	\begin{subfigure}[t]{0.196\linewidth}
		\centering
		\includegraphics[width=1\linewidth]{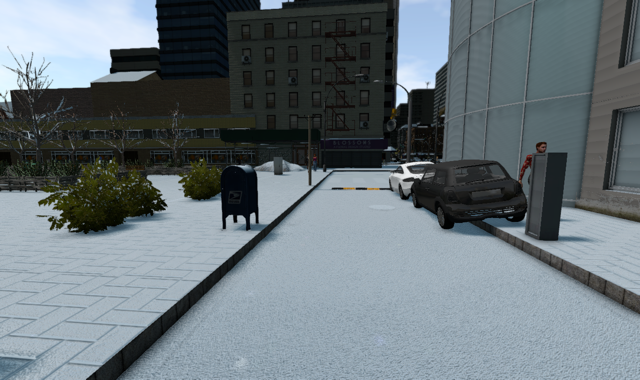}
	\end{subfigure}
	\hfill
	\begin{subfigure}[t]{0.196\linewidth}
		\centering
		\includegraphics[width=1\linewidth]{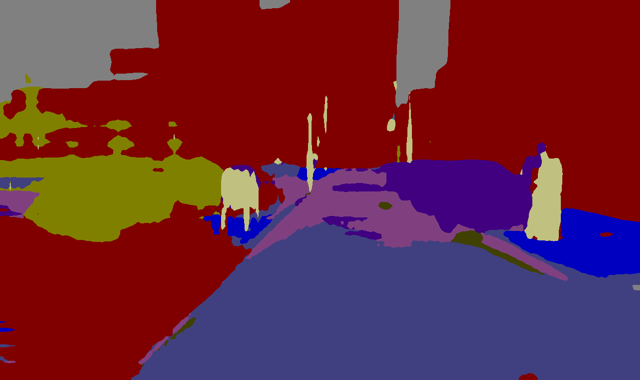}
	\end{subfigure}
	\hfill
	\begin{subfigure}[t]{0.196\linewidth}
		\centering
		\includegraphics[width=1\linewidth]{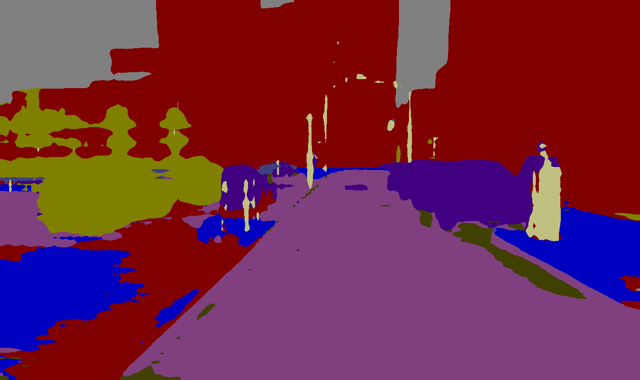}
	\end{subfigure}
	\hfill
	\begin{subfigure}[t]{0.196\linewidth}
		\centering
		\includegraphics[width=1\linewidth]{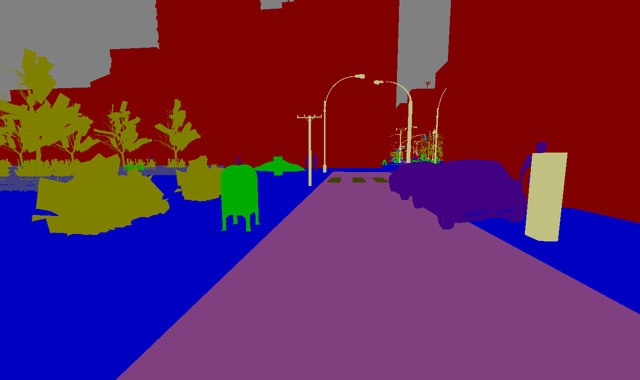}
	\end{subfigure}
	\hfill
	\\
	\centering
	\begin{subfigure}[t]{0.196\linewidth}
		\centering
		\includegraphics[width=1\linewidth]{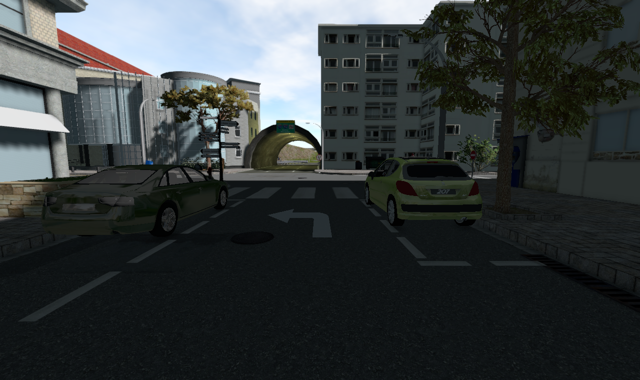}
		\caption{Fall Image}
	\end{subfigure}
	\hfill
	\begin{subfigure}[t]{0.196\linewidth}
		\centering
		\includegraphics[width=1\linewidth]{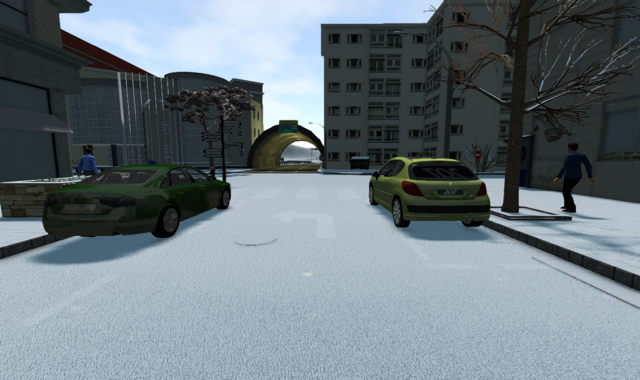}
		\caption{Winter Image}
	\end{subfigure}
	\hfill
	\begin{subfigure}[t]{0.196\linewidth}
		\centering
		\includegraphics[width=1\linewidth]{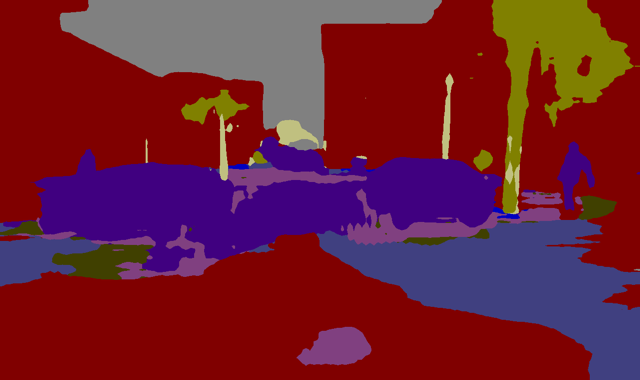}
		\caption{Before Adaptation}
	\end{subfigure}
	\hfill
	\begin{subfigure}[t]{0.196\linewidth}
		\centering
		\includegraphics[width=1\linewidth]{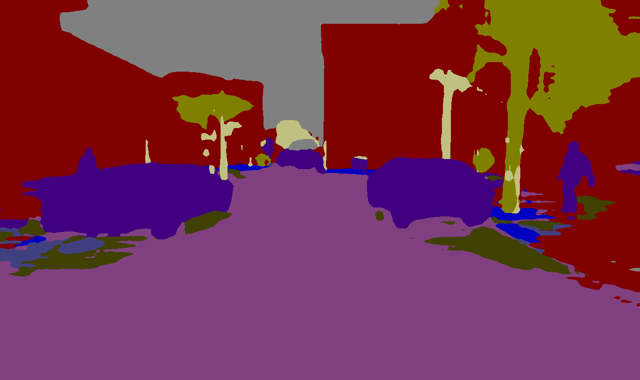}
		\caption{After Adaptation}
	\end{subfigure}
	\hfill
	\begin{subfigure}[t]{0.196\linewidth}
		\centering
		\includegraphics[width=1\linewidth]{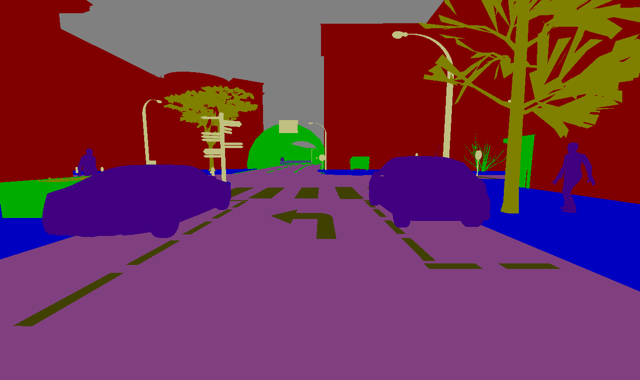}
		\caption{Ground Truth}
	\end{subfigure}
	\hfill
	\caption{Qualitative results on adaptation from cities in SYNTHIA \textit{fall} to cities in SYNTHIA \textit{winter}.}
	\label{fig:season2season}
    \vspace{-1mm}
\end{figure*}

%% file: table_season2season.tex
\begin{table*}[ht]
	\small
	\setlength{\tabcolsep}{3.9 pt}
	\begin{center}
		\begin{tabular}{l c c | ccccccccccccc g}
			\toprule
			Method & Source & Target (test) & \tblheaderSYNTHIA \\
			\midrule
			Before Adapt & Summer & Fall & 94.7 & 91.6 & 95.2 & 90.4 & 86.9 & 71.6 & 50.0 & 87.0 & 52.9 & 64.3 & 57.8 & 72.3 & 18.8 & 71.8 \\
			After Adapt & Summer & Fall & 95.0 & 91.6 & 94.6 & 90.3 & 86.8 & 71.7 & 49.4 & 87.3 & 54.7 & 64.3 & 54.2 & 69.7 & 20.8 & 71.6 \\ \hline
			Before Adapt & Fall & Summer & 95.0 & 92.9 & 96.7 & 91.9 & 90.2 & 71.8 & 53.4 & 93.2 & 50.6 & 62.3 & 48.1 & 82.7 & 14.7 & 72.8 \\ 
			After Adapt & Fall & Summer & 95.4 & 93.5 & 97.2 & 92.9 & 91.6 & 73.7 & 56.7 & 92.3 & 57.4 & 68.5 & 55.4 & 85.3 & 31.7 & 76.4 \\ \hline
			Before Adapt & Summer & Winter & 91.4 & 72.5 & 80.1 & 8.6 & 66.1 & 51.6 & 43.1 & 62.9 & 55.5 & 53.9 & 39.4 & 47.8 & 22.3 & 53.5 \\ 
			After Adapt & Summer & Winter & 90.7 & 71.1 & 78.4 & 9.2 & 64.0 & 50.9 & 42.8 & 60.5 & 56.1 & 53.5 & 40.7 & 49.2 & 23.0 & 53.1 \\ \hline
			Before Adapt & Winter & Summer & 91.2 & 90.5 & 82.5 & 34.2 & 53.3 & 59.1 & 49.2 & 85.7 & 44.7 & 62.8 & 46.3 & 44.6 & 28.9 & 59.5 \\ 
			After Adapt & Winter & Summer & 94.8 & 90.4 & 81.8 & 46.0 & 59.6 & 65.1 & 51.8 & 87.2 & 48.4 & 62.3 & 47.9 & 42.0 & 35.1 & 62.5 \\ \hline
			Before Adapt & Fall & Winter & 92.0 & 78.8 & 81.8 & 15.5 & 31.7 & 52.3 & 43.4 & 63.8 & 41.3 & 57.8 & 48.6 & 56.7 & 11.3 & 51.9 \\ 
			After Adapt & Fall & Winter & 92.1 & 86.7 & 91.3 & 20.8 & 72.7 & 52.9 & 46.5 & 64.3 & 50.0 & 59.5 & 54.6 & 57.5 & 26.1 & 59.6 \\ \hline
			Before Adapt & Winter & Fall & 94.3 & 88.0 & 85.5 & 32.0 & 56.2 & 60.9 & 48.7 & 77.2 & 47.0 & 57.9 & 49.8 & 46.6 & 27.1 & 59.3 \\ 
			After Adapt & Winter & Fall & 94.5 & 87.0 & 84.0 & 46.5 & 62.7 & 65.8 & 51.0 & 68.1 & 55.7 & 58.5 & 53.9 & 48.3 & 30.8 & 62.0 \\ \hline
			\rowcolor{Gray} Before Adapt & Avg & Avg & 93.1 & 85.7 & 87.0 & 45.4 & 64.1 & 61.2 & 48.0 & \bf 78.3 & 48.7 & 59.8 & 48.3 & 58.5 & 20.5 & 61.5 \\
            \rowcolor{Gray} After Adapt & Avg & Avg &  \bf 93.8 &  \bf 86.7 & \bf 87.9 & \bf  51.0 & \bf  72.9 & \bf  63.4 & \bf  49.7 & 76.6 & \bf  53.7 & \bf  61.1 & \bf  51.1 & \bf  58.7 & \bf  27.9 & \bf 64.2 \\ 
			\bottomrule
		\end{tabular}
	\end{center}
	\vspace{-3mm}
	\caption{\textbf{Adaptation across seasons.} We study the cross season performance using sub-sequences of SYNTHIA dataset. We report quantitative comparisons of performance before and after adaptation for training on one season and evaluating on another unannotated novel season. (Avg: the average performance of adaptation from one to another.)} 
	\label{table:season2season}
    \vspace{-5mm}
\end{table*}

%% file: table_city2city.tex
\begin{table*}[t]
\small
\setlength{\tabcolsep}{1.8pt}
\begin{center}
	\begin{tabular}{l | ccccccccccccccccccc  g}
		\toprule
		\multicolumn{21}{c}{Cityscapes \textit{train} $\rightarrow$ Cityscapes \textit{val}}\\
		\midrule
		Method & \tblheaderShort \\
		\midrule
		Dialation Frontend~\cite{yu2016multi} & 96.2 & 76.0 & 88.4 & 32.5 & 46.4 & 53.5 & 52.0 & 68.7 & 88.6 & 46.6 & 91.0 & 74.8 & 46.0 & 90.5 & 46.9 & 58.0 & 44.7 & 45.2 & 70.3 & 64.0 \\ 
		Our Method (GA Only) & \bf 97.0 & \bf 79.6 & 89.6 & \bf 42.8 & \bf 49.9 & 55.0 & 55.2 & \bf 70.2 & \bf 91.2 & \bf 59.8 & 92.5 & 75.4 & 46.5 & 91.6 & \bf 51.4 & \bf 66.0 & 49.3 & \bf 48.9 & \bf 71.6 & 67.6 \\ 
        Our Method (GA + CA) & \bf 97.0 & \bf 79.6 & \bf 89.8 & 42.2& 49.0& \bf 55.4 & \bf 56.3 & 70.1 & \bf 91.2 & \bf 59.8 &  \bf 92.6 & \bf 75.5 & \bf 48.1 & \bf 91.7 & 50.4 & 65.8 & \bf 53.2 & 48.0 & \bf 71.6 & \bf 67.8\\
		\bottomrule
	\end{tabular}
\end{center}
\vspace{-5mm}
\caption{\textbf{Adaptation across cities.} We study the performance using Cityscapes \textit{train} cities as source labeled training data adapted and evaluate our adaptation algorithm on Cityscapes \textit{val} as unlabeled target domains. Meanwhile, we show an ablation of the components of our method and how each contributes to the overall performance of our approach. Here GA indicates global domain alignment in section \ref{sec:global} and CA represents category specific adaptation in section \ref{sec:category}.}
\label{table:city2city}
\end{table*}

%% file: fig_cityscapes2nexar.tex

\begin{figure*}[t!]
	\centering
	\begin{subfigure}[t]{0.33\linewidth}
		\centering
		\includegraphics[width=1\linewidth]{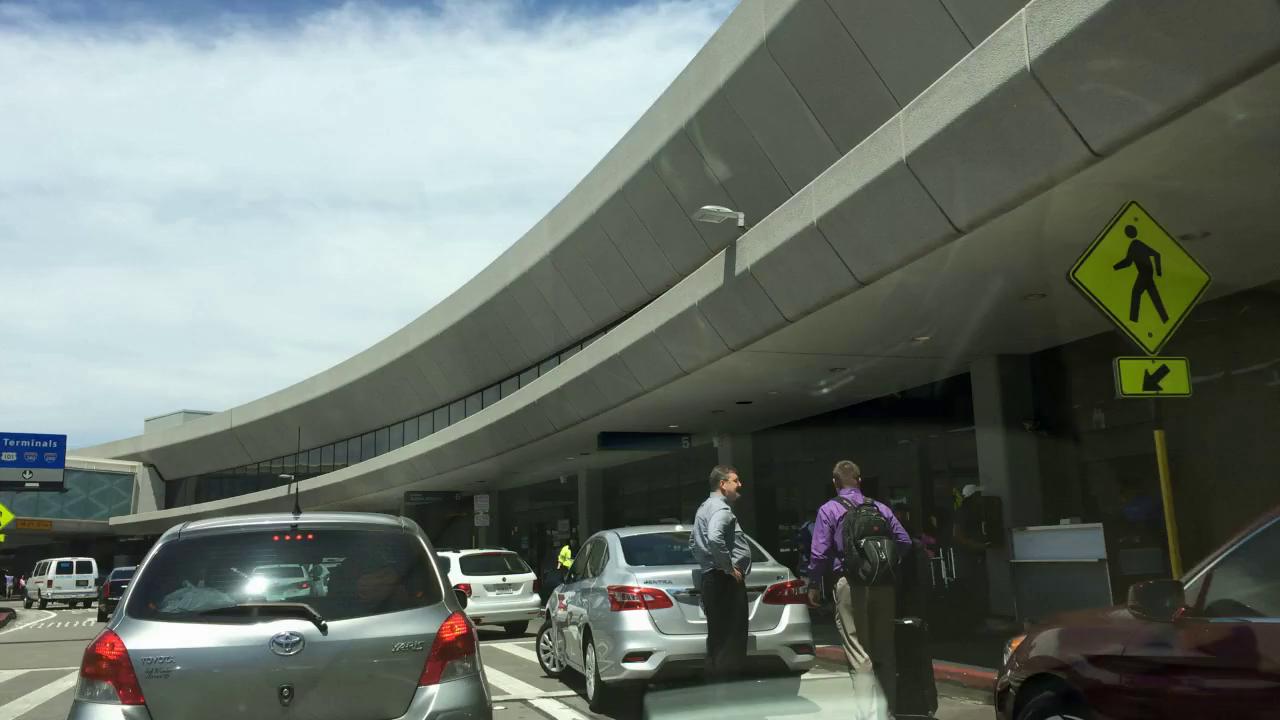}
	\end{subfigure}
	\hfill
	\begin{subfigure}[t]{0.33\linewidth}
		\centering
		\includegraphics[width=1\linewidth]{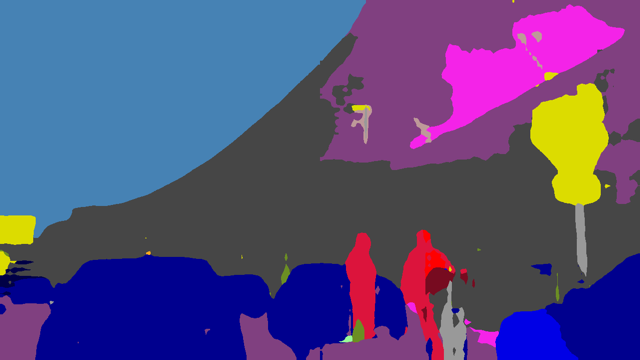}
	\end{subfigure}
	\hfill
	\begin{subfigure}[t]{0.33\linewidth}
		\centering
		\includegraphics[width=1\linewidth]{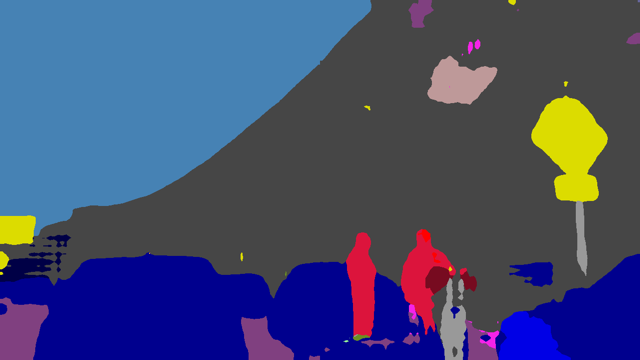}
	\end{subfigure}
	\hfill
	\\
	\centering
	\begin{subfigure}[t]{0.33\linewidth}
		\centering
		\includegraphics[width=1\linewidth]{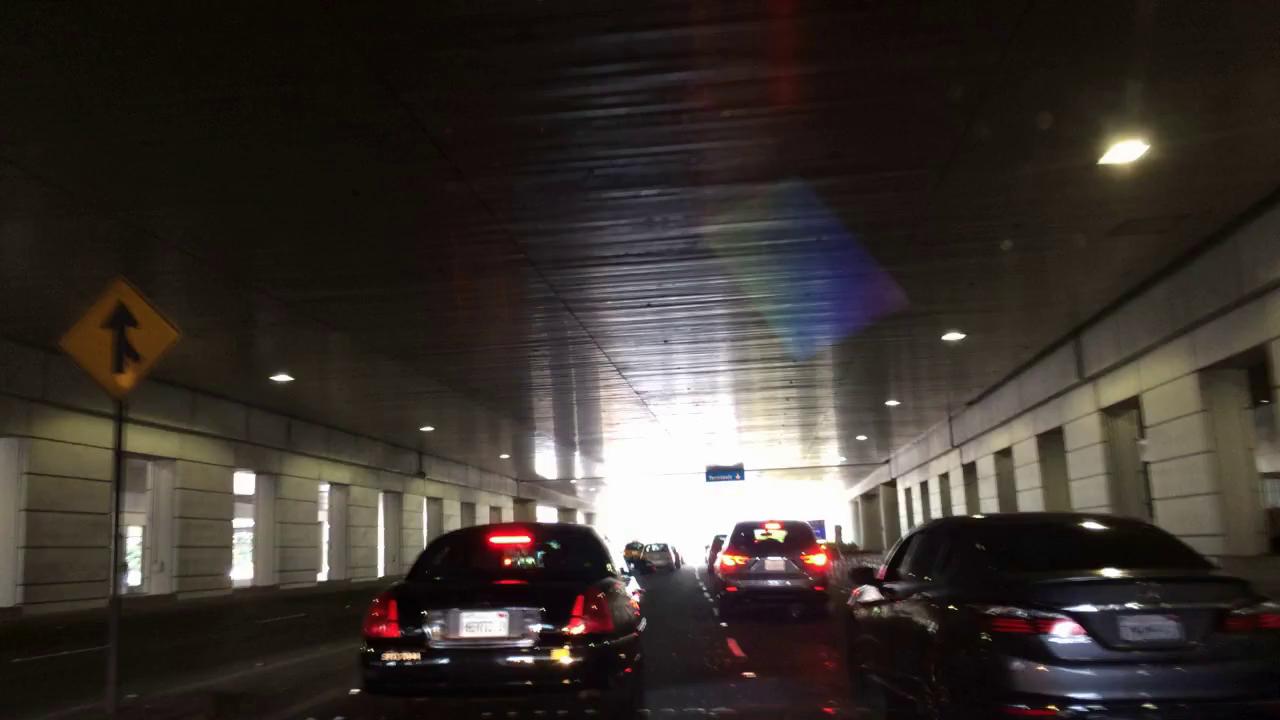}
        \caption{Original Image}
	\end{subfigure}
	\hfill
	\begin{subfigure}[t]{0.33\linewidth}
		\centering
		\includegraphics[width=1\linewidth]{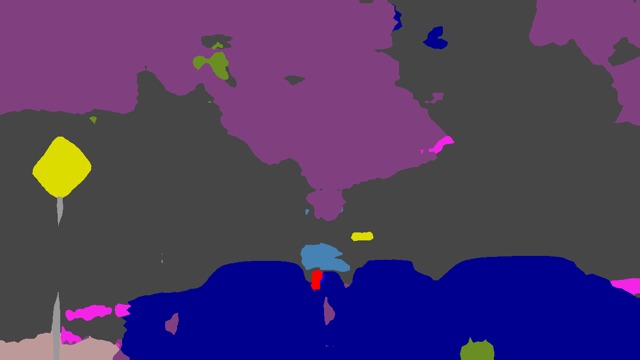}
        \caption{Before Adaptation}
	\end{subfigure}
	\hfill
	\begin{subfigure}[t]{0.33\linewidth}
		\centering
		\includegraphics[width=1\linewidth]{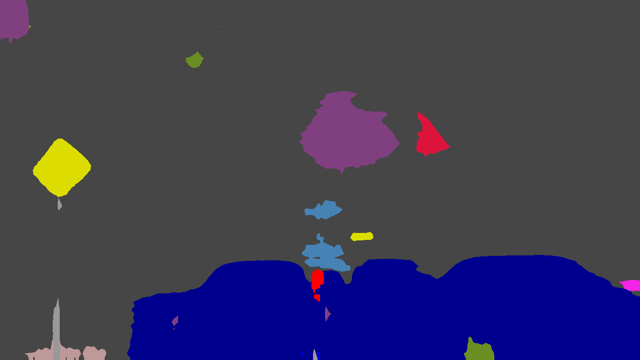}
        \caption{After Adaptation}
	\end{subfigure}
	\hfill
	\\
	\caption{Qualitative results on adaptation from cities in Cityscapes to cities in BDDS.}
	\label{fig:city2nexar}
\end{figure*}

%% file: conclusion.tex
\section{Conclusion}
\label{sec:conclusion}

In this paper, we present an unsupervised domain adaptation framework with fully convolutional networks for semantic segmentation. We propose fully convolutional networks with domain adversarial training for global domain alignment, while leveraging class-aware constrained multiple instance loss for transferring spatial layout. We demonstrate the effectiveness of our method on domain shifts between different cities, seasons and from synthetic to real, and we offer a new large-scale real-city driving image dataset. While the task of image classification has seen the bulk of the effort in developing domain adaptation methods, our experiments demonstrate the importance of adaptation in pixel-level dense prediction as well. Our approach is the first step in this direction.